\documentclass{article}

 \usepackage[nonatbib, final]{nips_2017}

\usepackage[utf8]{inputenc} 
\usepackage[T1]{fontenc}    
\usepackage{hyperref}       
\usepackage{url}            
\usepackage{booktabs}       
\usepackage{amsfonts}       
\usepackage{nicefrac}       
\usepackage{microtype}      
\usepackage{subfigure, graphicx, amsmath, algorithm, algorithmicx, algpseudocode, multirow}

\title{Image Segmentation with Pseudo-marginal MCMC Sampling and Nonparametric Shape Priors}

\author{
  Ertunc Erdil$^{1, 2}$\thanks{This work has been supported by the Scientific and Technological Research Council of Turkey (TUBITAK) under Grant 113E603.} \quad Sinan Y{\i}ld{\i}r{\i}m $^{1}$ \quad Tolga Tasdizen $^{3}$ \quad M\"{u}jdat \c{C}etin $^{1, 4}$\\
  $^1$ Faculty of Engineering and Natural Sciences,  Sabanci University, Istanbul, Turkey \\
  $^2$ ARM Ltd., 1 Summerpool Road, Loughborough, Leicester, UK \\
  $^{3}$ Department of Electrical and Computer Engineering, University of Utah, Salt Lake City, UT, USA \\
  $^{4}$ Department of Electrical and Computer Engineering, University of Rochester, Rochester, NY, USA \\
  \texttt{\{ertuncerdil, sinanyildirim, mcetin\}@sabanciuniv.edu} \quad \texttt{tolga@sci.utah.edu}
}

\begin{document}

\maketitle

\begin{abstract}
In this paper, we propose an efficient pseudo-marginal Markov chain Monte Carlo (MCMC) sampling approach to draw samples from posterior shape distributions for image segmentation. The computation time of the proposed approach is independent from the size of the training set used to learn the shape prior distribution nonparametrically. Therefore, it scales well for very large data sets. Our approach is able to characterize the posterior probability density in the space of shapes through its samples, and to return multiple solutions, potentially from different modes of a multimodal probability density, which would be encountered, e.g., in segmenting objects from multiple shape classes. Experimental results demonstrate the potential of the proposed approach.
\end{abstract}

\section{Introduction}
Incorporating prior shape density into the segmentation process has been widely studied in the literature \cite{kass1988snakes}, \cite{van2002active}, \cite{milborrow2008locating}, \cite{de2003adapting}, \cite{wang2002improved}, \cite{heimann2009statistical}. These methods can usually handle Gaussian-like, unimodal shape prior densities. Kim et al.~\cite{kim2007nonparametric} and Cremers et al.~\cite{cremers2006kernel} incorporate nonparametric density estimation based shape priors into the segmentation process using level sets. Therefore, these methods and their variants can learn ``multimodal" shape densities, which can be encountered in problems involving shape densities containing multiple classes of shapes \cite{foulonneau2009multi}, \cite{chen2009level}, \cite{yang2013shape}, \cite{souganli2014combining}, \cite{cremers2007review}, \cite{mesadi2017disjunctive}, \cite{erdil2017nonparametric}, \cite{8051060}. These methods minimize an energy function and find a solution at a local optimum. This does not provide any measure of the degree of confidence/uncertainty in that result and any information about the characteristics of the posterior density.

There are a limited number of Markov chain Monte Carlo (MCMC) based image segmentation methods in the literature. Most of these methods generate samples from the posterior density by assuming the prior density is uniform \cite{fan2007mcmc}, \cite{chang2011efficient}, \cite{chang2012efficient}. The only sampling-based segmentation approach in the literature that uses a shape prior is the one proposed by Chen et al.~\cite{chen2009markov}. However, the method cannot handle topological changes in shapes.

Our major contribution in this paper is a pseudo-marginal Markov chain Monte Carlo (MCMC) sampling-based image segmentation approach that exploits nonparametric shape priors. The proposed approach is able to characterize the posterior density through its samples. Our approach can learn from very large data sets efficiently by using pseudo-marginal sampling. To the best of our knowledge, this is the first approach that performs pseudo-marginal MCMC shape sampling-based image segmentation through an energy functional that uses nonparametric shape priors and level sets. Also, unlike other MCMC sampling-based segmentation approaches in the literature, the proposed approach perfectly satisfies the necessary conditions to implement MCMC sampling which is a crucial step for developing an MCMC sampler. 

A precursor to this work was presented in~\cite{erdil2016mcmc}. This paper advances that prior work in two major ways: (1) while~\cite{erdil2016mcmc} approximately satisfies the necessary conditions of MCMC sampling, the approach presented in this paper perfectly satisfies these conditions; (2) unlike~\cite{erdil2016mcmc} we use pseudo-marginal sampling to be able to learn shape densities from very large data sets; the method in~\cite{erdil2016mcmc} becomes inefficient with large training data.

\section{Model and problem definition} \label{sec: Model and problem definition}
The image segmentation problem involves estimating an unknown segmenting curve for an object given an observed image $y \in \mathcal{Y}^{M \times N}$ where $\mathcal{Y}$ is the set of the values that the pixels of $y$ can take. We are particularly interested in problems in which the prior shape distribution has components corresponding to different object classes. We denote the class of the object by $s \in \{ 1, \ldots, n \}$ where $n \geq 1$ is the total number of classes, which is known. For simplicity we assume that $s$ has a uniform distribution over $\{1,\ldots, n\}$ so that $p(s) = 1/n, \quad s  = 1, \ldots, n$.

We ultimately aim to estimate a binary segmenting curve $c \in \{ 0, 1\}^{M \times N}$ where $0$'s indicate the background and $1$'s indicate the object. The conditional density of $y$ given $c$ is independent from $s$ and is denoted by $p_{Y | C}(y | c)$. We construct this density based on the piecewise-constant version of the Mumford-Shah functional~\cite{mumford1989optimal}, \cite{chan2001active} which is a very common data fidelity term for image segmentation. To present the curves, we use level sets, which we define as a mapping $\phi: \{ 0, 1\}^{M \times N} \rightarrow \mathbb{R}^{M N}$. Our choice of level sets is to handle topological changes and use gradient flows effectively in our methodology. We denote the level set of $c$ as $x = \phi(c)$. 

We also have a training set of binary curves that are grouped into classes, $\mathbf{C} = \{ \mathcal{C}_{1}, \ldots, \mathcal{C}_{n} \}$, where each $\mathcal{C}_{i} = \{ c_{i, 1}, \ldots, c_{i, m_{i}} \}$ is the collection of $m_{i} \geq 1$ segmented curves for class $i$. Let us also define the level set representation of the training set as $\mathbf{X} = \{ \mathcal{X}_{1}, \ldots, \mathcal{X}_{n} \}$, where each $\mathcal{X}_{i} = \{ x_{i, 1}, \ldots, x_{i, m_{i}} \}$ with $x_{i, j} = \phi(c_{i, j})$, the level set representation of $c_{i, j}$. Now we can define the prior distribution for $x$ given the class $s$ as $p(x | s) = \frac{1}{m_{s}} \sum_{i = 1}^{m_{s}} \mathcal{N} (x; x_{s, i}, \sigma^{2} I)$, where $\mathcal{N} (x; \mu, \Sigma)$ is a Gaussian density with mean $\mu$ and covariance matrix $\Sigma$. This prior corresponds to a mixture of kernels with centers $x_{s, 1}, \ldots, x_{s, m_{s}}$ with kernel size $\sigma$ \cite{kim2007nonparametric}. To determine the kernel size $\sigma$, we use an ML kernel with leave-one-out~\cite{silverman1986density}. Let us also define $\bar{\phi}$ as the pseudo-inverse\footnote{Note that $\phi$ is not invertible, so we define a pseudo-inverse.} of $\phi$ such that $\bar{\phi} (\phi(c)) = c$. We use $\bar{\phi}$ to rewrite the conditional density of the data in terms of $x$ as $p(y | x) = p_{Y | C}(y | \bar{\phi}(x))$. We can also write the joint density of $s$, $x$, and $y$ as $p(s, x, y) = p(s) p(x | s) p(y | x)$.

The Bayesian image segmentation problem can be formulated as finding the posterior distribution of $x$ given $y$ which is $p(x | y) \propto p(y | x)  p(x) =p(y | x) \sum_{s = 1}^{n} p(s) p(x | s)$. However, estimating $p(x | y)$ can be difficult since the summation over classes makes the distribution hard to infer, e.g., using Monte Carlo sampling methods. Alternatively, we aim for the joint posterior distribution of $s$ and $x$ given $y$, $p(s, x | y) \propto p(s, x, y)$, whose marginal is still the desired posterior $p(x | y)$.

\section{Methodology} \label{sec: Methodology}
\subsection{Metropolis-Hastings within Gibbs} \label{sec: Metropolis-Hastings with Gibbs}
We aim to sample from $p(s, x | y)$ using Gibbs sampling by sampling from the conditional densities $p(s | y, x)$ and $p(x | y, s)$ in an alternating fashion \cite{geman1984stochastic}, \cite{gelfand1990sampling}. However, since the full conditional $p(x | s, y)$ is  hard to sample from, we update $x$ by using a Metropolis-Hastings (MH) move, which leads to the well known Metropolis-Hastings within Gibbs (MHwG) algorithm \cite{geweke2001bayesian}.

Both conditional densities in MHwG involve $p(x | s)$ which needs to be evaluated during MH updates. This can be too costly when $m_{s}$ is large, which occurs when we have a big training set. Therefore, towards a more computationally efficient MCMC algorithm that scales with the training data size,  we consider the following unbiased estimator of $p(x | s)$ via subsampling: $\widehat{p}(x | s) = \frac{1}{\hat{m}_{s}} \sum_{j = 1}^{\hat{m}_{s}} \mathcal{N} (x; x_{s, u_{j}}, \sigma^{2} I)$ where $\{ u_{1}, u_{2}, \ldots, u_{\hat{m}_{s}}\} \subset \{1, 2, \ldots, m_{s} \}$ is a set of subsamples generated via sampling without replacement and $\hat{m}_{s} \ll m_{s}$. This approximation of the prior leads to the approximation of the conditional posterior densities $\widehat{p}(s | x, y) \propto p(s) \widehat{p}(x | s)$ and $\widehat{p}(x | s, y) \propto \widehat{p}(x | s) p(y | x)$.
Using this approximation does not generally guarantee that the Markov Chain has an equilibrium distribution that is exactly $p(x, s|y)$. To achieve a correct MH algorithm which uses the approximation $\widehat{p}(x | s)$ and still targets $p(x, s | y)$, we adopt the pseudo-marginal MH algorithm of \cite{andrieu2009pseudo}; the details are described in the following section.

\subsection{Pseudo-marginal MHwG}
Assume that we have a non-negative random variable $z$ such that given $x$ and $s$, its conditional density given $s$ and $x$, $g_{s, x}(z)$, satisfies $\int_{0}^{\infty} g_{s, x}(z) z dz = p(x | s)$. We choose $z$ as an approximation to  $p(x | s)$, in particular $z = \widehat{p}(x | s)$ where $\widehat{p}(x | s)$ is defined in Section \ref{sec: Metropolis-Hastings with Gibbs}, and its probability density $g_{s, x}(z)$ corresponds to the generation process of this approximation. (It will become clear that in fact we do not have to calculate $g_{s, x}(z)$ at all but we should be able to sample from it.) Note that $z$ is a random variable since we generate $\{u_1, \dots, u_{\hat{m}_s}\}$ randomly when computing $\widehat{p}(x | s)$. We can define the extended posterior density with the new variable $z$ added as $p(x, s, z | y) \propto p(s) z g_{s, x}(z) p(y | x)$. When we integrate $z$ out, we see that samples for $s$ and $x$ from $p(x, s, z | y)$ will admit the desired posterior $p(s, x | y)$: $p(s) p(y | x) \int z g_{s, x}(z) dz = p(s) p(x | s) p(y | x)$. Now, the problem of generating samples from $p(s, x | y)$ can be replaced with the problem of generating samples from $p(x, s, z | y)$.

We propose a pseudo-marginal MHwG sampling procedure to generate samples from $p(x, s, z | y)$. Note the important remark that this algorithm also targets $p(x, s | y)$, hence $p(x | y)$ exactly. There are two major steps of the proposed approach which we explain in detail in the following: (1) condition the posterior on $x$ and update $s$ and $z$, (2) update $x$ and $z$ by conditioning the posterior on $s$.

\noindent \textbf{Update step for the class $s$ and $z$:} The distribution from which we sample $s$ and $z$ in Metropolis-Hastings is $p(s, z|y, x)$. We can write this distribution as $p(s, z | y, x) = p(s, z | x) \propto p(s)z g_{s, x}(z)$. Since we regard the proposal mechanism as a joint update of $s$ and $z$, the proposal generates $(s', z')$ from the density $q(s' | s)g_{s', x}(z)$. Note that $(s', z')$ denotes the candidate samples generated from the proposal distribution. We take $q(s' | s)$ as a uniform distribution $\mathcal{U}\{1, n\}$ and $z' = \widehat{p}(x |s')$ \cite{andrieu2009pseudo}. Once $s'$ and $z'$ are sampled, they are either accepted with probability
\begin{equation}
\min \left\{ 1, \frac{p(s') z' g_{s', x}(z') q(s | s') g_{s, x}(z)}{p(s) z g_{s, x}(z) q(s' | s) g_{s', x}(z')} \right\}
= \min \left\{ 1, \frac{p(s') z' q(s | s') }{p(s) z q(s' | s) } \right\},
\label{eq:MHratioClass}
\end{equation}
or the current values of $s$ and $z$ are kept. The probabilities in the MH ratio in \eqref{eq:MHratioClass} can be computed exactly which guarantees satisfying the necessary conditions to implement MCMC sampling.

\noindent \textbf{Update step for the level set $x$ and $z$:} The next step is to sample $x$ and $z$ from the conditional density $p(x, z|y, s)$. To achieve this, we perform Metropolis-Hastings sampling as we use for sampling $s'$ and $z'$. The conditional density $p(x, z|y, s)$ can be written as $p(x, z | s, y) \propto z g_{s, x}(z) p(y | x)$. Also, for joint sampling of candidates $(x', z')$ we can write the proposal density as $q_{s, j}(x', z' | y) = q_{s, j}(x' | x, y) g_{s, x'}(z')$ where $j$ is sampled uniformly from $\{1, m_{s}\}$, and $z' = \widehat{p}(x' | s)$ is generated using subsampling. Then, the Metropolis-Hastings ratio can be computed as follows:
\begin{equation}
\min \left\{ 1, \frac{z' g_{s', x}(z') p(y | x') q_{s, j}(x | x', y) g_{s, x}(z)}{z g_{s, x}(z) p(y | x) q_{s, j}(x' | x, y) g_{s', x}(z')} \right\} = \min \left\{ 1, \frac{z' p(y | x') q_{s, j}(x | x', y)}{z p(y | x) q_{s, j}(x' | x, y)} \right\}
\label{eq:MH_shape}
\end{equation}

\noindent \textbf{Design of the proposal distribution:} The crucial part in \eqref{eq:MH_shape} is designing the proposal distribution to generate a candidate curve $x'$ from $x$. Let us define an energy function whose gradient drives the curve to the desired location using the data and the training images in a particular class $s$ as $E_{s}(x) := \log p(x | s) + \log p( y | x, s)$. When the training data set is too large, calculating $\log p(x | s)$ may be too expensive as discussed earlier. An unbiased estimator of $E_{s}(x)$ would be obtained as $E_{s, j}(x) := \log \mathcal{N} (x; x_{s, j}, \sigma^{2} I) + \log p( y | x, s)$ where $j \sim \mathcal{U}(1, \ldots, m_{s})$. The proposal distribution after sampling the $j^{th}$ training image in class $s$ is then constructed as $q_{s, j}(x' | x, y) = \mathcal{N} \left(x'; x - \widehat{\nabla E}_{s, j}(x), \Sigma \right)$. Here, the shift term $\widehat{\nabla E}_{s, j}(x)$ is an approximation\footnote{Note that the term $(y - \mu_{\textup{in}})^{\odot 2} - (y - \mu_{\textup{out}})^{\odot 2} $ is a discrete approximation w.r.t.\ the level set representation $x$ \cite{chan2001active}, so that the whole expression is an approximation to $\nabla E_{s, j}(x)$.} to the gradient $\nabla E_{s, j}(x)$ w.r.t.\ $x$ and is given by $\widehat{\nabla E}_{s, j}(x) = \frac{1}{\sigma^{2}} (x - x_{s, j})  + \left[ (y- \mu_{\textup{in}})^{\odot 2} - (y - \mu_{\textup{out}})^{\odot 2} \right]$ where $(\cdot)^{\odot k}$ is the element-wise power operation.

In the design of $q_{s, j}(x' | x, y)$, the most important part is selecting a covariance matrix, $\Sigma$, that generates smooth perturbations since smoother curves are more likely~\cite{kim2007nonparametric}, \cite{fan2007mcmc}. In the proposed approach, we compute a positive semi-definite covariance matrix $\Sigma$ such that it generates smooth random perturbations. Given an $M \times N$ image, we first generate an $M N \times M N$ matrix $Z$ by drawing i.i.d.\ samples from a unit Gaussian distribution. Then, we construct another $MN \times MN$ matrix $F$ where each column of $F$ is a smoothed version of each row in $Z$. By assuming $F$ is constructed by multiplying $Z$ by a matrix $\widehat{A}$, we can find the matrix $\widehat{A}$ as $\widehat{A} = Z^{-1}F$ since $Z$ is generally invertible. Given $\widehat{A}$, a covariance matrix $\widehat{\Sigma}$ can be computed by $\widehat{\Sigma} = \widehat{A}\widehat{A}^T$. However, generally $\widehat{\Sigma}$ is not positive semi-definite since $\widehat{A}$ is not generally a full rank matrix. Therefore, we find the closest positive semi-definite matrix to $\widehat{\Sigma}$ using the approach in~\cite{highamNearest} which we take it as $\Sigma$.

\section{Experimental results}
We compare running time of the proposed pseudo-marginal sampling approach and conventional sampling approach which uses all training examples to estimate the prior shape density. We perform experiments on the MNIST data set \cite{lecun1998gradient} which contains 60,000 training examples. We construct training sets with various sizes from $1K$ to $50K$ by randomly selecting equal number of samples from each digit class. We generate $1000$ samples using both the pseudo-marginal sampling and conventional sampling on a test image by using training sets with different sizes. The plot in Figure \ref{fig:time_sample_size} shows average running time as a function of training set size for both pseudo-marginal sampling and conventional MCMC sampling. The average single sample generation time of the proposed approach does not change as the training set size increases since we choose $\hat{m}_{s} = 10$ in all experiments. We also measure the segmentation accuracy of all samples with the ground truth using Dice score \cite{dice1945measures}. Average Dice score\footnote{Note that Dice score takes values in $[0, 1]$ where $1$ indicates the perfect match with the ground truth.} result for pseudo-marginal sampling is $0.7718$ whereas it is $0.7777$ for the conventional sampling. The very slight decrease in Dice results of pseudo-marginal sampling can be acceptable in many applications when the huge gain in running time is considered.

    \begin{minipage}{.50\textwidth}
\begin{figure}[H]
\centering
\includegraphics[scale=0.18]{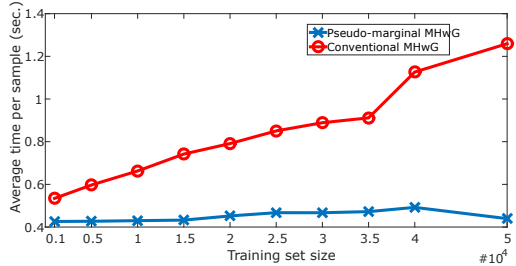}
\caption{Average running time for producing a single sample as a function of training set size.\label{fig:time_sample_size}}
\end{figure}
    \end{minipage}\hfill
\begin{minipage}{.47\textwidth}
\begin{figure}[H]
\centering
\subfigure[Test image\label{fig:MNIST_speed_test_segmentation}]{\includegraphics[scale=0.5]{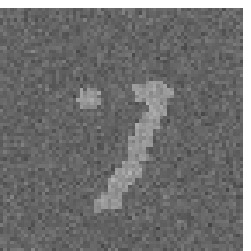}}
\subfigure[Kim et al. \cite{kim2007nonparametric}
\label{fig:MNIST_kimetal_mcb}]{\includegraphics[scale=0.5]{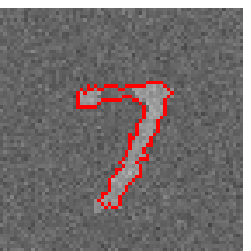}}
\subfigure[The proposed approach\label{fig:MNIST_multimodal3_mcb}]{\includegraphics[scale=0.5]{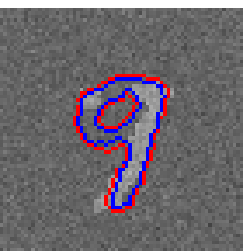}
\includegraphics[scale=0.5]{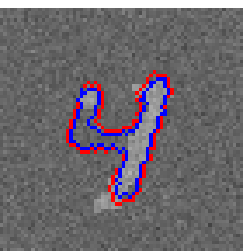}
\includegraphics[scale=0.5]{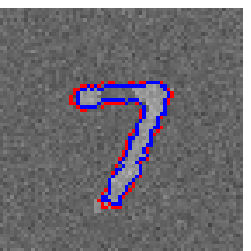}}
\caption{Marginal confidence bounds obtained by samples on a test image. Note that in (c) red indicates the least and blue indicates the most confidence boundaries.\label{fig:MNIST_multimodal_mcb}}
\end{figure}
    \end{minipage}

We also run the proposed approach on the test image in Figure \ref{fig:MNIST_speed_test_segmentation} and generate $1000$ samples. The algorithm generates samples from digit classes $4, 7,$ and $9$ as shown in Figure \ref{fig:MNIST_multimodal3_mcb}. The optimization-based approach of Kim et al. \cite{kim2007nonparametric} finds a single segmentation solution shown in Figure \ref{fig:MNIST_kimetal_mcb}. This experiment shows that our algorithm can produce samples from different modes.

\section{Conclusion}
We propose a pseudo-marginal Markov chain Monte Carlo (MCMC) sampling-based image segmentation approach that exploits nonparametric shape priors. The proposed approach generates samples from the posterior distribution $p(x | y)$ to avoid shortcomings of the optimization-based approaches which include getting stuck at local optima and being unable to characterize the posterior density. The proposed MCMC sampling approach deals with all these problem while being computationally efficient, unlike the conventional MCMC approaches, by using pseudo-marginal sampling principles. Moreover, our pseudo-marginal shape sampler perfectly satisfies the necessary conditions to implement MCMC sampling which is crucial for ensuring the generated samples come from the desired distribution. Existing methods in the literature only approximately satisfy these conditions.


\clearpage
\newpage
\bibliographystyle{spmpsci}      
\bibliography{refs} 

%
%
%
%

\end{document}